\pdfoutput=1

\documentclass[11pt]{article}

\usepackage{svg}

\usepackage[final]{acl}

\usepackage{times}
\usepackage{latexsym}

\usepackage[ruled, lined, linesnumbered, commentsnumbered, longend]{algorithm2e}
\usepackage{algpseudocode}
\usepackage{multirow}

\usepackage{color}
\usepackage{makecell}
\usepackage{upgreek}
\usepackage[dvipsnames]{xcolor}
\usepackage{tikz}

\usepackage[T1]{fontenc}
\usepackage[utf8]{inputenc}

\usepackage{microtype}

\usepackage{inconsolata}

\usepackage{graphicx}

\title{$\mathbf{SearchLLM}$: Detecting LLM Paraphrased Text by Measuring the Similarity with Regeneration of the Candidate Source via Search Engine}

\author{Hoang-Quoc Nguyen-Son$^\dagger$, Minh-Son Dao$^\dagger$, and Koji Zettsu $^{\dagger\ddagger}$  \\
  $^\dagger$National Institute of Information and Communications Technology, Japan \\
  \tt \{quoc-nguyen,dao,zettsu\}@nict.go.jp \\ 
  $^\ddagger$ Nagoya University, Japan \\
  \tt zettsu@i.nagoya-u.ac.jp}

\begin{document}
\maketitle
\begin{abstract}
With the advent of large language models (LLMs), it has become common practice for users to draft text and utilize LLMs to enhance its quality through paraphrasing. However, this process can sometimes result in the loss or distortion of the original intended meaning. Due to the human-like quality of LLM-generated text, traditional detection methods often fail, particularly when text is paraphrased to closely mimic original content. In response to these challenges, we propose a novel approach named $\mathrm{SearchLLM}$, designed to identify LLM-paraphrased text by leveraging search engine capabilities to locate potential original text sources. By analyzing similarities between the input and regenerated versions of candidate sources, $\mathrm{SearchLLM}$ effectively distinguishes LLM-paraphrased content. $\mathrm{SearchLLM}$ is designed as a proxy layer, allowing seamless integration with existing detectors to enhance their performance. Experimental results across various LLMs demonstrate that $\mathrm{SearchLLM}$ consistently enhances the accuracy of recent detectors in detecting LLM-paraphrased text that closely mimics original content. Furthermore, $\mathrm{SearchLLM}$ also helps the detectors prevent paraphrasing attacks.
\end{abstract}

\section{Introduction}

\begin{figure}[!t]
    \centering
    \includegraphics[width=1\linewidth]{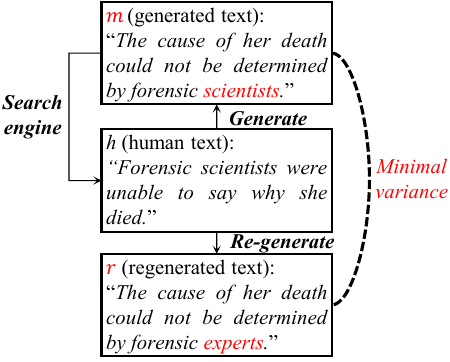}
    \caption{Illustration of variance in LLM-generated texts derived from a human text $h$. The comparison demonstrates minimal variance between the generated text $m$ and the regenerated text $r$.}
    \label{fig:motivation}
\end{figure}

Large language models (LLMs) have demonstrated remarkable capabilities in supporting a wide range of linguistics-related tasks. Despite their power and versatility, the increasing reliance on LLMs also raises concerns regarding potential misuse, particularly when users employ these models to paraphrase texts without adequately reviewing the generated outputs. 
This practice can result in the final text failing to accurately convey the original intent, potentially leading to misunderstandings among recipients. As a result, there is an urgent need for robust detection mechanisms to identify and differentiate LLM-paraphrased content from genuine human writing. 

Exisiting LLM detectors can be categorized into three primary approaches.
The watermarking approach involves embedding specific characteristics within the generated text, allowing for differentiation between human-authored and LLM-generated content~\cite{kirchenbauer2023watermark,chang2024postmark,gloaguen2025black}. However, this method faces limitations with closed LLM models where such control is not feasible. The second approach leverages supervised learning, utilizing extensive datasets of both human and LLM-generated texts to train models that can distinguish between the two~\cite{hu2023radar,li2024mage,xu2024detecting}. While effective, this method is sensitive to out-of-distribution scenarios, which can compromise its reliability. The third approach employs zero-shot techniques, enabling detection without prior training by exploiting inherent differences in text structure and style~\cite{mitchell2023detectgpt,hans2024spotting,park2025dart}. 

Despite these varied strategies, detecting LLM-generated text that closely mimics human writing remains a formidable challenge. Consequently, we propose focusing on zero-shot methods to enhance current detection systems, aiming to address the nuanced problem of identifying text that seamlessly integrates into human-like narratives. This proposal builds on existing frameworks and seeks to fortify the detection capabilities against increasingly sophisticated LLM outputs.

\textbf{Motivation}: In our investigation, we observe a notable consistency in the outputs generated by a specific LLM across multiple trials. For instance, a human-authored text, denoted as $h$, was randomly selected from the XSum dataset~\cite{narayan2018don}, and GPT-4o-mini was utilized with default temperature settings to produce a paraphrased version, $m$, as shown in Figure~\ref{fig:motivation} with the prompt adapt from \citet{zhu2023beat}: ``\textit{Paraphrase the following text: <human text>.}'' Subsequently, employing the same model, a re-generated text, $r$, was created from $h$. The striking similarity between $m$ and $r$ suggests that if the original human text $h$ can be retrieved,  via search engine capabilities, it becomes feasible to detect the paraphrased output $m$. 

\begin{figure}[!t]
    \centering
    \includegraphics[width=1\linewidth]{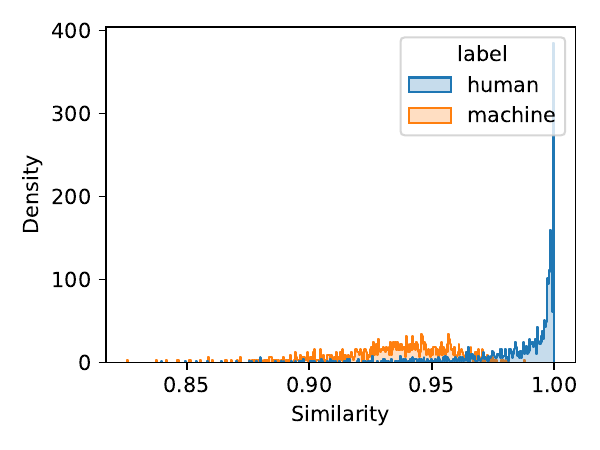}
    \caption{Comparison of similarity between samples produced by humans and LLMs with internet sources.}
    \label{fig:input_similarity}
\end{figure}

\begin{figure}[!t]
    \centering
    \includegraphics[width=1\linewidth]{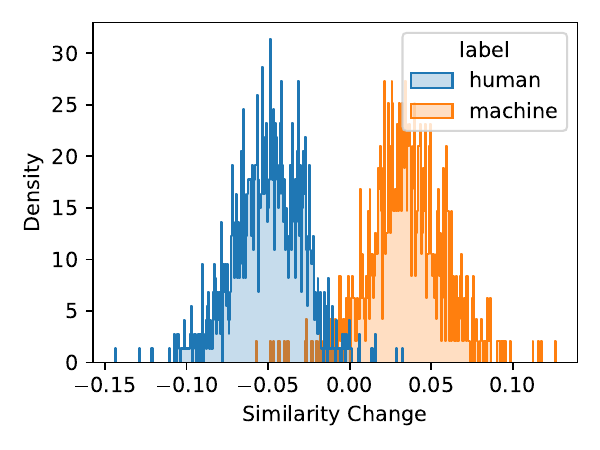}
    \caption{Change in similarity between regenerated samples and human or LLM-generated samples.}
    \label{fig:regeneration_similarity}
\end{figure}

We compare the similarity between 1,000 human-written samples and their corresponding sources retrieved from the internet, as well as 1,000 LLM-paraphrased samples and their sources, as shown in Figure~\ref{fig:input_similarity}.
The results indicate that human-written samples generally exhibit higher similarity to their sources than the LLM-paraphrased samples. However, a few human-written samples show low similarity, likely due to noise in the internet sources, such as updates since their initial creation or issues with inconsistent parsing by crawling tools. To further investigate these cases, we regenerate both the human and LLM samples using GPT-4o-mini and analyze the change in similarity between the regenerated and human/LLM samples, as shown in Figure~\ref{fig:regeneration_similarity}. Since GPT-4o-mini tends to regenerate LLM-paraphrased text that is more similar to the internet source, the similarity shift in these cases is generally positive. In contrast, when GPT-4o-mini regenerates human-written text that diverges from the internet source, the similarity shift tends to be negative.

\textbf{Contribution}: We introduced a method called $\mathrm{SearchLLM}$ to identify text paraphrased by LLMs. The method operates by initially utilizing search engines to identify candidate original texts corresponding to the input text. Subsequently, $\mathrm{SearchLLM}$ produces re-generated texts from these candidate originals. By comparing the input text with the re-generated versions, $\mathrm{SearchLLM}$ can effectively discern whether the text is authored by a human or paraphrased by an LLM.
Our contribution can be summarized as follows:

\begin{itemize}
    \item We propose a method called $\mathrm{SearchLLM}$, which can detect text paraphrased by LLMs.

    \item We design $\mathrm{SearchLLM}$ to function as a proxy, allowing it to enhance the performance of any existing detectors\footnote{Source code is available at \url{https://github.com/quocnsh/SearchLLM}}.

    \item We conducted experiments with various datasets, which demonstrate that $\mathrm{SearchLLM}$ effectively improves the performance of existing detectors. Additionally, $\mathrm{SearchLLM}$ efficiently helps these detectors identify text that closely mimics original content and are consistent with paraphrasing attacks. 
\end{itemize}

\section{Related Work}
\label{sec:appendix:related_work}

The LLM text detection methods can be categorized into three main approaches: watermarking, supervised-based, and zero-shot. 

\textbf{Watermarking}: An LLM is guided to produce outputs that contain specific, embedded characteristics~\cite{gloaguen2025black, lu2024entropy, huo2024token, chang2024postmark}. These characteristics act as subtle signals which can help distinguish LLM-generated text from human-authored content. For instance, \citet{kirchenbauer2023watermark} demonstrated a method where the LLM is instructed to preferentially use terms from a “green” word list while avoiding those from a “red” word list, thereby embedding a detectable linguistic signature within the output. Such watermarking not only enables highly effective detection of LLM-generated text, but also provides verifiable evidence to support the identification process.

Widespread adoption of techniques to safeguard media authenticity is evident, such as the Coalition for Content Provenance and Authenticity (C2PA) initiative used to watermark AI-generated images like those from GPT Image. However, watermarking text is more challenging due to the context-dependent nature of language and requires direct control over the LLM output, limiting it to closed models. Additionally, watermarking carries risks like piggyback spoofing, where attackers embed watermark patterns into human-written text, undermining detection reliability.

\textbf{Supervised-Based}: 
The prevailing methods entail collecting large corpora and training a classifier on the final layer of transformer-based models. Moreover, recent findings have demonstrated that leveraging representations from intermediate layers can yield additional benefits for distinguishing subtle textual characteristics~\cite{yu2024text}. Complementary to this, \citet{verma2024ghostbuster} expand the feature set by integrating $n$-gram statistics, which enhances the model’s expressiveness and robustness. Furthermore, \citet{tian2024multiscale} introduce a multiscale positive-unlabeled detection framework tailored for separately processing short texts, addressing length-based variability in classification performance. 

To further improve classification accuracy, some techniques involve generating augmented argument samples through paraphrasing~\cite{hu2023radar}, thereby enriching the diversity of training data. Contract learning has also emerged as a promising strategy, offering a systematic means to explicitly distinguish between outputs generated by LLMs and those produced by humans~\cite{liu2024does,guo2024detective,liu2023coco}. In addition, the adoption of specialized techniques, such as syntactic parsing trees~\cite{park2025dart,li2025prdetect,kim2024threads} and Fourier transformation~\cite{xu2024detecting}, facilitates the extraction of structural and frequency-based features, deepening the model’s analytical capacity. 

\textbf{Zero-Shot}: Another prominent approach for detecting LLM-generated text is zero-shot learning, which sidesteps the need for dedicated training by directly leveraging statistical or behavioral properties of LLM outputs. A notable method in this category was proposed by \citet{mitchell2023detectgpt}, who demonstrated that LLM-generated text tends to be optimized for higher probability under the generation model compared to human-written text. Building on this foundational insight, subsequent methods have sought to improve detection by enhancing either speed~\cite{bao2024fast} or detection accuracy~\cite{su2023detectllm,zeng2024dald,xu2025training}. Some techniques further adapt this probabilistic hypothesis to develop black-box detectors by relying on an alternative model~\cite{mireshghallah2024smaller,ma2024zero,shi2024ten,bao2025glimpse,wu2023llmdet} or two alternative models~\cite{hans2024spotting}. Additionally, \citet{guo2024biscope} exploit the LLM’s ability to memorize previously generated words to distinguish machine-generated content, whereas other methods~\cite{koike2024outfox,bhattacharjee2024fighting} directly task an LLM with self-identification of its own outputs. 

Closely related to our hypothesis, recent approaches recognize that LLMs often produce text with distinct generative patterns, prompting the use of the LLM itself to regenerate a text and compare similarity to the input text~\cite{zhu2023beat,nguyen2024simllm,mao2024raidar,yang2024dna,yu2024dpic}. The key distinction from our method is that, while prior work typically employs the input text for regeneration, we draw upon candidate source materials retrieved from the internet as the basis for text regeneration. Nonetheless, both supervised and zero-shot detection techniques are notably sensitive to out-of-distribution texts and are vulnerable to attacks, such as paraphrasing~\cite{krishna2023paraphrasing}, which significantly undermine their robustness in adversarial scenarios.

\section{$\mathbf{SearchLLM}$}
\textbf{Overview}: Figure~\ref{fig:LLM_text_detection_overview} illustrates the workflow employed by $\mathrm{SearchLLM}$ for detecting text generated by LLMs. Given an input text $t$, the process starts by utilizing a search engine to retrieve a candidate source text $t_c$ that is related to $t$ and could plausibly serve as its origin. Next, the similarity score $\sigma_c$ between the original candidate text $t_c$ and the input text $t$ is computed. If $\sigma_c$ exceeds a predetermined threshold $\alpha$, the input text $t$ is classified as human-written, under the assumption that significant textual overlap with verifiable sources indicates human authorship. If $\sigma_c$ does not surpass the threshold, the method proceeds by leveraging the LLM to re-generate a new text $t_r$ based on the candidate $t_c$. The similarity between $t_r$ and the original input $t$ is then calculated; a sufficiently high similarity score at this stage suggests that $t$ was likely generated by an LLM. In cases where neither condition is conclusive, the detection of $t$ is deferred to an existing detection method. The details of the process are described in Algorithm~\ref{alg:LLM_text_detection} of Appendix~\ref{appendix:session:algorithm}. The main steps of $\mathrm{SearchLLM}$ for detecting LLM-generated text are summarized as follows:

\begin{figure*}[!t]
    \centering
    \includegraphics[width=1\linewidth]{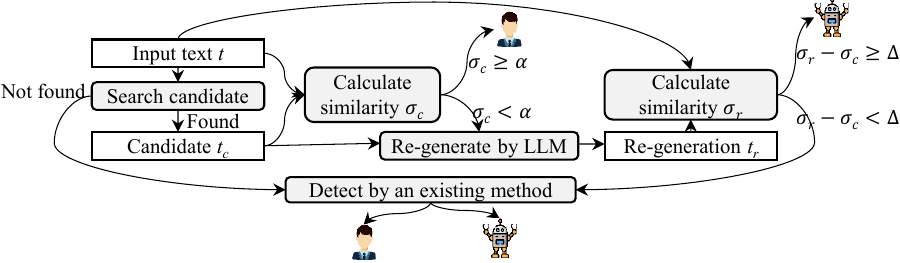}
    \caption{ Overview of the $\mathrm{SearchLLM}$ schema for determining whether input text $t$ is human- or LLM-generated. The system considers three main cases: (1) $\mathrm{SearchLLM}$ compares $t$ with a candidate $t_c$ retrieved by a search engine to identify human text; (2) $\mathrm{SearchLLM}$ generates a regeneration $t_r$ from $t_c$ to determine LLM-generated text; (3) if neither case applies, $\mathrm{SearchLLM}$ delegates the decision to an existing method.}
    \label{fig:LLM_text_detection_overview}
\end{figure*}

\begin{figure}[!t]
    \centering
    \includegraphics[width=1\linewidth]{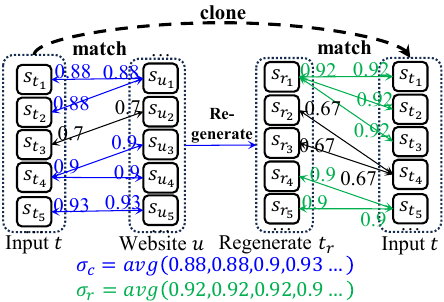}
    \caption{Process of matching between input data and website content or re-generated text.}
    \label{fig:matching_sample}
\end{figure}

\textbf{Candidate Extraction}:
This step assesses whether the text at a given URL $u$ could serve as a potential input source for generating the target text $t$. We observe that input text may be rewritten from the original source. After rewriting, some sentences may be merged or separated. Therefore, we propose a greedy matching approach between the input text and the retrieved source, as illustrated in Figure~\ref{fig:matching_sample}.

Both $t$ and the text from $u$ are segmented into individual sentences.
First, we identify the sentence $s_{u_1}$ from $u$ that has the highest similarity to the first sentence $s_{t_1}$ in the input text, serving as an anchor. Similarity is measured using a fixed pretrained model\footnote{\url{https://huggingface.co/Qwen/Qwen3-Embedding-0.6B}}. Starting from the anchor, we find the optimal matching between the first sentences in $u$ and the first sentence $s_{t_1}$. We also compare the optimal matching between the first sentences in $t$ and $s_{u_1}$. The best matching is retained. This process is repeated until all sentences in $t$ are matched.
The matching process is demonstrated in Algorithm~\ref{alg:matching} of Appendix~\ref{appendix:session:algorithm}.

\textbf{Similarity Measurement}:
We estimate similarity based on the matching method described above. Specifically, we evaluate sentences with similarity scores greater than the machine threshold $\beta$, which is fixed at 0.86\footnote{All hyperparameters are determined via beam search with a step size of 0.01 and are held constant for all experiments in this paper.}. If the number of remaining sentences exceeds the ratio threshold $\gamma$ at 0.5 and their average similarity is greater than the human threshold $\alpha$ (set at 0.99), we confidently classify $t$ as human-written. Otherwise, we use the prompt ``\textit{Paraphrase the following text: <candidate text $t_c$>}'' to generate a regenerated version of the text, $t_r$, based on matched sentences in $u$, and calculate a new similarity score $\sigma_r$ between $t$ and $t_r$ using the filtered sentences. If the increase from $\sigma_c$ to $\sigma_r$ exceeds the predefined threshold $\Delta$ (where $\Delta = 0.01$), we classify $t$ as LLM-generated. If neither condition is met, the input $t$ is subsequently passed to an existing detection method for further analysis.

\section{Evaluation}

\subsection{General Scenario}

We conduct experiments on the RAID dataset~\cite{dugan2024raid}, which contains human-written and LLM-generated texts. We utilize all available Wikipedia-related texts and employ the Wikipedia search engine for the $\mathrm{SearchLLM}$. The LLM text is generated using 11 LLMs from five  providers: GPT, LLaMA, Cohere, MPT, and Mistral. Specifically, we use representative models from each provider, including GPT-4, Llama-2-70B-Chat, Cohere-Chat, MPT-30B-Chat, and Mistral-7B-Chat. This process yields a total of 10,794 samples, which are equally distributed between human-authored content and text generated by these five representative LLMs. 

We use GPT-4o-mini within $\mathrm{SearchLLM}$ to create regenerated texts by paraphrasing. Since the LLM-generated texts in this dataset are produced using topic-based prompts, their meanings often differ from those of human-written texts. Therefore, any LLM used in $\mathrm{SearchLLM}$ to create regenerated texts tends to yield same performance.

We evaluate a wide range of representative methods, including $\mathrm{RADAR}$~\cite{hu2023radar}, $\mathrm{Longformer}$~\cite{li2024mage}, $\mathrm{DetectGPT}$~\cite{mitchell2023detectgpt}, and $\mathrm{Binoculars}$~\cite{hans2024spotting}. Watermarking-based detectors are not applicable to GPT-4 in the RAID dataset. $\mathrm{Longformer}$ is a supervised detector trained on a large volume of text from the MAGE dataset~\cite{li2024mage}. $\mathrm{RADAR}$ uses the Vicuna-7B model, which is trained on an argumentation dataset with paraphrased texts via adversarial learning. $\mathrm{DetectGPT}$ estimates changes in probability on the GPT-2-XL model after perturbations generated by the T5-large model. $\mathrm{Binoculars}$ measures perplexity using two models: Falcon-7B and Falcon-7B-Instruct.
For evaluation, we adopt a ROC AUC metric and ROC AUC at false positive rate's of 1\%, standard approaches widely used for assessing the performance of LLM detectors. The $F$-score is  shown in Appendix~\ref{appendix:other_metrics}.

\begin{table*}[!t]
    \centering
    \setlength\tabcolsep{6.0pt} 
    \begin{tabular}{l c c c c c }  
         \textbf{Method}&  \textbf{GPT-4}&  \textbf{Llama-2-70B }&  \textbf{Cohere }&  \textbf{MPT-30B }& \textbf{Mistral-7B }\\ \hline 
         $\mathbf{Longformer}$&  0.9932&  0.9970&  0.9505&  0.9916& 0.9895\\ 
 $\mathbf{SearchLLM}$& \textbf{0.9979}& \textbf{0.9996}&\textbf{0.9794}& \textbf{0.9975}&\textbf{0.9966}\\\hline
 $\mathbf{RADAR}$& 0.9906
& 0.9968& 0.9599& 0.9962&0.9947\\
 $\mathbf{SearchLLM}$& \textbf{0.9949}& \textbf{0.9982}& \textbf{0.9793}& \textbf{0.9979}&\textbf{0.9970}\\\hline
 $\mathbf{Detect}$& 0.9022& 0.9167& 0.7629& 0.8542&0.8868\\
 $\mathbf{SearchLLM}$& \textbf{0.9379}&\textbf{0.9546}&  \textbf{0.8793}& \textbf{0.9233}& \textbf{0.9399}\\\hline
 $\mathbf{Binoculars}$
& 0.9987& \textbf{0.9999}& 0.9977&\textbf{0.9999}&0.9998\\
 $\mathbf{SearchLLM}$& \textbf{0.9994}&\textbf{0.9999}& \textbf{0.9989}& \textbf{0.9999}& \textbf{0.9999}\\

    \end{tabular}
    \caption{Performance of LLM-generated text detection on all Wikipedia-related samples from the RAID dataset (ROC AUC).}
    \label{tab:RAID_wikipedia_ROC_AUC}
\end{table*}

\begin{table*}[!t]
    \centering
    \setlength\tabcolsep{6.0pt} 
    \begin{tabular}{l c c c c c l}  
         \textbf{Method}&  \textbf{GPT-4}&  \textbf{Llama-2-70B }&  \textbf{Cohere }&  \textbf{MPT-30B }& \textbf{Mistral-7B }&\textbf{Average}\\ \hline 
         $\mathbf{Longformer}$&  0.4824&  0.4820
&  0.2505&  0.3924
& 0.4008 &0.4016\\ 
 $\mathbf{SearchLLM}$& \textbf{0.8912}& \textbf{0.9297}&\textbf{0.6102}& \textbf{0.7857}
&\textbf{0.8456} &\textbf{0.8124}\\\hline
 $\mathbf{RADAR}$& 0.6190& 0.7493& 0.4241& 0.7628
&0.7003 &0.6511
\\
 $\mathbf{SearchLLM}$& \textbf{0.7166}& \textbf{0.8268}& \textbf{0.4863}& \textbf{0.8235}&\textbf{0.7758} &\textbf{0.7258}
\\\hline
 $\mathbf{DetectGPT}$& 0.0174& 0.0340& 0.0097& 0.0186& 0.0292
&
0.0218
\\
 $\mathbf{SearchLLM}$& \textbf{0.0639}&\textbf{0.0896}& \textbf{0.0288}& \textbf{0.0502}&  \textbf{0.0822}&
\textbf{0.0629}\\\hline
 $\mathbf{Binoculars}$
& \textbf{0.7736}& 0.0561& \textbf{0.9233}&0.0561&0.5046 &0.4627\\
 $\mathbf{SearchLLM}$& 0.7207&\textbf{0.6182}& 0.8754& \textbf{0.6182}& \textbf{0.6732} &\textbf{0.7011}\\

    \end{tabular}
    \caption{Performance of LLM-generated text detection on all Wikipedia-related samples from the RAID dataset (ROC AUC at an FPR of 1\%).}
    \label{tab:RAID_wikipedia_ROC_AUC_FPT_1}
\end{table*}

Table~\ref{tab:RAID_wikipedia_ROC_AUC} shows that the existing detectors are effective at identifying LLM-generated text across all LLM models. Among the supervised-based detectors, $\mathrm{Longformer}$ and $\mathrm{RADAR}$ achieve nearly identical performance on each model. In contrast, among the zero-shot-based detectors, $\mathrm{DetectGPT}$ performs slightly worse than the supervised detectors, while $\mathrm{Binoculars}$ outperforms all existing detectors with  a ROC AUC exceeding 0.997. $\mathrm{SearchLLM}$ further enhances the performance of all detectors. Specifically, it maintains $\mathrm{Binoculars}$’ performance at 0.9999 on LLaMa and MPT models and improves the performance in other cases. In Table~\ref{tab:RAID_wikipedia_ROC_AUC_FPT_1}, $\mathrm{SearchLLM}$ achieves competitive ROC AUC at false positive rate's of 1\% compared to $\mathrm{Binoculars}$ and outperforms the other methods.

\begin{figure}[!t]
    \centering
    \includegraphics[width=1\linewidth]{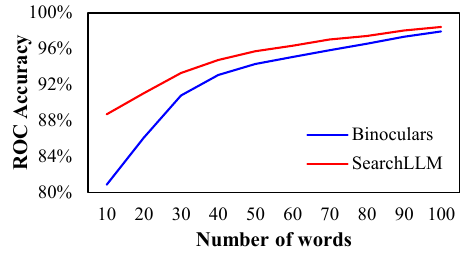}
    \caption{Comparison of $\mathrm{Binoculars}$ and $\mathrm{SearchLLM}$ in detecting LLM-generated text with limited words.}
    \label{fig:text_length}
\end{figure}

\textbf{Short Text}:
We conduct experiments using short texts, with lengths limited to between 10 and 100 words, as shown in Figure~\ref{fig:text_length}. 
The experiments are performed with GPT-4 text and $\mathrm{SearchLLM}$ on the representative method, $\mathrm{Binoculars}$. The results demonstrate that $\mathrm{SearchLLM}$ can improve the performance of $\mathrm{Binoculars}$, especially with short texts. In particular, $\mathrm{Binoculars}$ combined with $\mathrm{SearchLLM}$ achieves a ROC AUC of 89.7\%, when the text length is only 10 words.

\subsection{Rigorous Scenario}

To rigorously assess the ability to detect LLM-generated text that closely resembles human-written content, we design an evaluation scenario in which various large language models (LLMs) are tasked with paraphrasing text originally authored by Wikipedia humans as illustrated in Table~\ref{tab:RAID_paraphrasing}. Specifically, we utilize a diverse selection of state-of-the-art LLMs, including GPT-4o-mini (4o-mini), GPT-4.1 (4.1), DeepSeek-V3-0324 (DeepSeek), Grok-3-mini (Grok), and Phi-4 (Phi). We use the corresponding LLM in $\mathrm{SearchLLM}$ to create the regenerated text. The scenarios involving unknown LLMs and prompts are further detailed later.

\begin{table}[!t]
    \centering
\setlength\tabcolsep{1.5pt} 
    \begin{tabular}{l c c c c c }  
         \textbf{Method}&    \textbf{4o-mini}
&\textbf{4.1}&  \textbf{DeepSeek}&  \textbf{Grok}& \textbf{Phi}\\ \hline 
         $\mathbf{Long}$&    0.6552
&0.6805&  0.4997&  0.4996&  0.5018\\   
 $\mathbf{Search}$&   \textbf{0.9073}
&\textbf{0.9173}& \textbf{0.8630}& \textbf{0.8607}& \textbf{0.8145}\\\hline
 $\mathbf{RA}$&   0.6543
&0.6436& 0.6070& 0.6038& 0.6023\\
 $\mathbf{Search}$&   \textbf{0.9062}
&\textbf{0.9107}& \textbf{0.8926}& \textbf{0.8906}& \textbf{0.8473}\\\hline
 $\mathbf{Detect}$&   
0.6073&0.5572& 0.5761& 0.5593& 0.5751\\
 $\mathbf{Search}$&   
\textbf{0.8876}&\textbf{0.8817}& \textbf{0.8792}& \textbf{0.8779}& \textbf{0.8675}\\\hline
 $\mathbf{Bino}$&   0.6388
&0.6127
& 0.4327& 0.4351& 0.4363\\
 $\mathbf{Search}$& \textbf{0.9070}  
&\textbf{0.9064}& \textbf{0.8481}&\textbf{0.8456}& \textbf{0.7919}\\

    \end{tabular}
    \caption{Detection of paraphrased text generated by various large language models.}.
    \label{tab:RAID_paraphrasing}
\end{table}

The results demonstrate that existing methods experience a significant decline in performance when applied to this rigorous scenario. Specifically, many of the results show a sharp drop in a ROC AUC to around 0.6. 
In contrast, paired $t$ tests conducted in Appendix~\ref{appendix:section:statisical_test} demonstrate that $\mathrm{SearchLLM}$ significantly improves the performance of all existing methods across various models, consistently maintaining a ROC AUC above 0.79 in every case.

\begin{figure}[!t]
    \centering
    \includegraphics[width=1\linewidth]{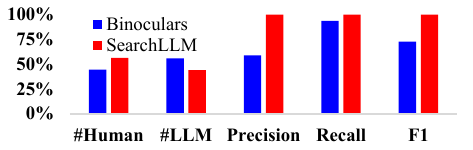}
    \caption{The number of human- and LLM-generated texts processed by $\mathrm{SearchLLM}$ and $\mathrm{Binoculars}$, along with their performance in detecting LLM-generated content.}
    \label{fig:breakdown_performance_gpt_4o_mini}
\end{figure}

\textbf{Breakdown of Performance}:
We present a breakdown of $\mathrm{SearchLLM}$'s performance in conjunction with $\mathrm{Binoculars}$ for detecting LLM-generated text by GPT-4o-mini, as shown in Figure~\ref{fig:breakdown_performance_gpt_4o_mini}.
When LLM-generated text is paraphrased, it becomes more challenging for the search engine to retrieve than human-written text. Consequently, $\mathrm{SearchLLM}$ processes 56.2\% of human text, and 43.8\% of LLM-generated text. Among the texts that are processed, $\mathrm{SearchLLM}$ achieves higher precision, recall, and $F1$ scores overall compared to $\mathrm{Binoculars}$. Similar results for detecting GPT-4.1-generated text are presented in Appendix~\ref{sec:appendix:breakdown_gpt_4_1}.

\textbf{Ablation Studies}: We evaluate the effect of each component of $\mathrm{SearchLLM}$ on detecting text generated by GPT-4o-mini, as shown in Table~\ref{tab:ablation_studies_components}. 
The three main components are: $\alpha$, for detecting human-written text; $\beta$, for detecting LLM-generated text; and $\Delta$, for rechecking the LLM-generated text. 
Since $\mathrm{SearchLLM}$ processes more human-written text than machine-generated text, $\alpha$ is more important than $\beta$. $\Delta$ also demonstrates the importance of rechecking LLM-generated text. Overall, all variants of $\mathrm{SearchLLM}$ still outperform the corresponding existing approaches that do not use $\mathrm{SearchLLM}$.

\begin{table}[!t]
    \centering
\setlength\tabcolsep{1pt} 
    \begin{tabular}{ l   c c c c}
\multirow{2}{*}{\textbf{Variant}}  & $\mathbf{Search}$& $\mathbf{Search}$& $\mathbf{Search}$&$\mathbf{Search}$\\
 & $\mathbf{(Long)}$& $\mathbf{(RA)}$& $\mathbf{(Detect)}$&$\mathbf{(Bino)}$\\ \hline 
 \textbf{w/o $\alpha$ }&  0.7920&  0.8068& 0.7376&0.7952\\   
 \textbf{w/o $\beta$} & 0.8466& 0.8466& 0.8021&0.8359\\ 
 \textbf{w/o $\Delta$} & 0.7914& 0.7865& 0.7561&0.7891\\ 
 \textbf{w/o $\mathbf{Search}$} & 0.6552& 0.6543& 0.6073
&0.6388\\ 
 \textbf{Full model} & \textbf{0.9073}& \textbf{0.9062}& \textbf{0.8876}&\textbf{0.9070}\\ 
    \end{tabular}
    \caption{Ablation studies examining the impact of different hyperparameter settings on $\mathrm{SearchLLM}$.}
    \label{tab:ablation_studies_components}
\end{table}

\textbf{Change of Parameters}: We analyze the impact of varying the human threshold $\alpha$ while keeping all other parameters constant as shown in Figure~\ref{fig:human_threshold_alpha}.
Other parameters are detailed in Appendix~\ref{sec:appendix:other_parameters}. When $\alpha$ is set below 0.92, $\mathrm{SearchLLM}$ achieves performance comparable to $\mathrm{Binoculars}$. As $\alpha$ increases beyond 0.96, $\mathrm{SearchLLM}$ attains more stable performance.

\begin{figure}[!t]
    \centering
    \includegraphics[width=1\linewidth]{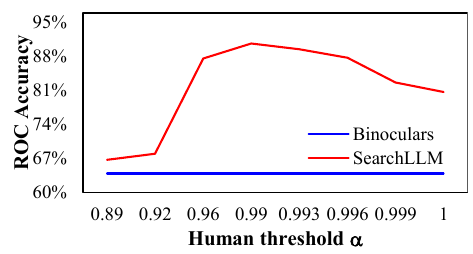}
    \caption{The impact of varying the human threshold parameter $\alpha$ on the performance of $\mathrm{SearchLLM}$.}
    \label{fig:human_threshold_alpha}
\end{figure}

\textbf{Unknown Models and Prompts}: We conduct experiments to detect text generated by large language models (LLMs) when the specific LLM and/or prompt is unknown, as summarized in Table~\ref{tab:unknown_model_prompt}. 
 Experiments involving unknown temperatures are described in Appendix~\ref{appendix:section:temperature}.
Specifically, we use GPT-4o-mini (4o-mini) to identify text produced by various LLMs, including GPT-4.1 (4.1), DeepSeek-V3 (DeepSeek), Grok-3-mini (Grok), and Phi-4 (Phi). We also evaluate detection performance on texts generated with different prompts, including versions that have been revised or polished. To create revised text, we use the prompt ``\textit{\underline{$\mathit{Revise}$} the following text: <human text>.}''; for polished text, we replace ``\textit{Revise}'' with ``\textit{Polish}'' in the prompt. The results demonstrates that $\mathrm{SearchLLM}$ still achieves the stable performances with unknown models and prompts.
$\mathrm{SearchLLM}$ achieves comparable results in detecting composite text, as discussed in Appendix~\ref{appendix:section:composite}.

\begin{table*}[!t]
    \centering
    \begin{tabular}{ l  c c c c c c c c}
\multirow{2}{*}{\textbf{Generation}} & $\mathbf{Para}$& $\mathbf{Para}$ & $\mathbf{Para}$ 
& $\mathbf{Para}$ 
&$\mathbf{Para}$ 
& $\mathbf{Revise}$ &$\mathbf{Revise}$ &$\mathbf{Polish}$\\
 & \textbf{4o-mini}& \textbf{4.1} & \textbf{DeepSeek}& \textbf{Grok}&\textbf{Phi}& \textbf{4o-mini} &\textbf{4.1} &\textbf{4o-mini}\\\hline
 \multirow{2}{*}{\textbf{Regeneration}}& $\mathbf{Para}$& $\mathbf{Para}$ & $\mathbf{Para}$ 
& $\mathbf{Para}$ 
&$\mathbf{Para}$ 
& $\mathbf{Para}$ &$\mathbf{Para}$ &$\mathbf{Para}$\\
 & \textbf{4o-mini}& \textbf{4o-mini} & \textbf{4o-mini} 
& \textbf{4o-mini} 
&\textbf{4o-mini} 
& \textbf{4o-mini} &\textbf{4o-mini} &\textbf{4o-mini}\\ \hline 
         $\mathbf{Longformer}$&  0.6552&   0.6805& 
0.4997& 
0.4996&
0.5018& 0.6693 &0.6761&0.6337\\   
 $\mathbf{SearchLLM}$& \textbf{0.9073}&  \textbf{0.8750}& \textbf{0.8332}& \textbf{0.8342}& \textbf{0.8318}& \textbf{0.8851} &\textbf{0.8528}&\textbf{0.8019}\\ \hline 
 $\mathbf{RADAR}$& 0.6543&  0.6436& 
0.6070& 
0.6038&
0.6023& 0.6849 &0.6761&0.6910\\ 
 $\mathbf{SearchLLM}$& \textbf{0.9062}&  \textbf{0.8540}& \textbf{0.8652}& \textbf{0.8646}&\textbf{0.8623}& \textbf{0.8876} &\textbf{0.8433}&\textbf{0.8148}\\\hline
 $\mathbf{DetectGPT}$& 0.6073
&  0.5572& 
0.5664& 
0.5593&
0.5751&  0.5993
&0.5826&0.6355\\
 $\mathbf{SearchLLM}$& \textbf{0.8876}&   \textbf{0.8396}& \textbf{0.8654}& \textbf{0.8475}& \textbf{0.8700}&  \textbf{0.8259}& \textbf{0.8492}&\textbf{0.7917}\\\hline
 $\mathbf{Binoculars}$
& 0.6388
& 0.6127& 
0.4327& 
0.4351&
0.4363&  0.6576 &0.6670&0.6523\\
 $\mathbf{SearchLLM}$& \textbf{0.9070}& \textbf{0.8478}& \textbf{0.8118}& \textbf{0.8112}&\textbf{0.8108}& \textbf{0.8800} &\textbf{0.8492}&\textbf{0.8027}\\
    \end{tabular}
    \caption{Detection of LLM-generated text when the LLMs and prompts are unknown.}
    \label{tab:unknown_model_prompt}
\end{table*}

\begin{table*}[!t]
    \centering
\setlength\tabcolsep{1.5pt} 
    \begin{tabular}{ l | c  c | c c c c }  
         \textbf{Dataset}&  \textbf{MAGE (News)}&\textbf{MAGE (QA)}&  \multicolumn{4}{c}{\textbf{XSum}}\\\hline
\multirow{2}{*}{\textbf{Generation}} & $\mathbf{Topic}$-$\mathbf{based}$ &$\mathbf{Topic}$-$\mathbf{based}$ & $\mathbf{Paraphrase}$ & $\mathbf{Paraphrase}$& $\mathbf{Revise}$&$\mathbf{Polish}$\\
 & \textbf{3.5-turbo} &\textbf{3.5-turbo} & \textbf{4o-mini}& \textbf{4o} & \textbf{4o-mini}&\textbf{4o-mini}\\
 \multirow{2}{*}{\textbf{Regeneration}}& $\mathbf{Paraphrase}$ &$\mathbf{Paraphrase}$ & $\mathbf{Paraphrase}$& $\mathbf{Paraphrase}$& $\mathbf{Paraphrase}$&$\mathbf{Paraphrase}$\\
 & \textbf{4o-mini} &\textbf{4o-mini} & \textbf{4o-mini}&  \textbf{4o-mini}& \textbf{4o-mini}&\textbf{4o-mini}\\ \hline 
         $\mathbf{Longformer}$& 0.9957&0.9998& 0.7179& 0.6749&0.7360&0.7440\\   
 $\mathbf{SearchLLM}$& \textbf{0.9957}&\textbf{1.000}& \textbf{0.8587}& \textbf{0.8257}&\textbf{0.8652}&\textbf{0.8486}\\ \hline 
 $\mathbf{RADAR}$& 0.9425&0.7992& 0.7554& 0.7069&0.7728&0.7665\\ 
 $\mathbf{SearchLLM}$& \textbf{0.9591}&\textbf{0.8654}& \textbf{0.8621}&  \textbf{0.8194}&\textbf{0.8759}&\textbf{0.8367}\\\hline
 $\mathbf{DetectGPT}$& 0.7079&0.9505& 0.4014& 0.4424&0.4262&0.4761\\
 $\mathbf{SearchLLM}$& \textbf{0.8064}&\textbf{0.9639}& \textbf{0.6884}& \textbf{0.6936}&\textbf{0.7016}&\textbf{0.6808}\\\hline
 $\mathbf{Binoculars}$
& 0.9992&0.9940& 0.5576& 0.5370&0.6112&0.6287\\
 $\mathbf{SearchLLM}$& \textbf{0.9995}&\textbf{0.9961}& \textbf{0.7701}& \textbf{0.7387}&\textbf{0.8063}&\textbf{0.7768}\\
    \end{tabular}
    \caption{Detection of LLM generated text using Google Search Engine.}
    \label{tab:google_search}
\end{table*}

\textbf{Running Time}: We evaluate the latency of the detectors as presented in Table~\ref{tab:running_time}. 
When combined with other methods, $\mathrm{SearchLLM}$ with $\mathrm{RADAR}$ achieves a similar runtime to $\mathrm{DetectGPT}$ and requires less than 10 seconds in the worst case.

\begin{table}[!t]
    \centering
        \setlength\tabcolsep{6.0pt} 
    \begin{tabular}{l c c}  
         \textbf{Method}&  \textbf{w/o $\mathbf{Search}$ }&\textbf{with $\mathbf{Search}$ }\\ \hline 
         $\mathbf{Longformer}$&  3.2s &7.3s 
\\ 
 $\mathbf{RADAR}$& 0.5s &5.9s 
\\
 $\mathbf{DetectGPT}$&7.4s &9.4s 
\\
 $\mathbf{Binoculars}$
&6.2s &8.8s \\

    \end{tabular}
    \caption{Comparison of running times for existing methods with and without $\mathrm{SearchLLM(Search)}$  support.}
    \label{tab:running_time}
\end{table}

\textbf{Attack Scenario}: We evaluate the resilience of the detectors under adversarial conditions by employing the $\mathrm{DIPPER}$ method~\cite{krishna2023paraphrasing} with default settings\footnote{\url{https://huggingface.co/kalpeshk2011/dipper-paraphraser-xxl}} 
(lexical diversity = $60$, order diversity = $0$, top $p$ = $0.75$) to generate attacks.
Specifically, we use $\mathrm{DIPPER}$ to target either the original human-written text, the LLM text paraphrased by GPT-4o-mini, or both. This approach allows us to systematically assess how well the detectors can maintain performance when confronted with deliberately altered input designed to evade detection. The empirical results, summarized in Table~\ref{tab:DIPPER_attack}, provide the robustness of the detectors against such adversarial manipulation.

\begin{table}[!t]
    \centering
        \setlength\tabcolsep{6.0pt} 
    \begin{tabular}{l c c c }  
         \textbf{Method}&  \textbf{Human} &  \textbf{LLM}&  \textbf{Both}\\ \hline 
         $\mathbf{Longformer}$&  0.5629&  0.6069&  0.5042\\   
 $\mathbf{SearchLLM}$& \textbf{0.7644}& \textbf{0.7896}& \textbf{0.5316}\\ \hline 
 $\mathbf{RADAR}$& 0.5306& 0.6484& 0.5309\\ 
 $\mathbf{SearchLLM}$& \textbf{0.7686}& \textbf{0.7939}& \textbf{0.5603}\\\hline
 $\mathbf{DetectGPT}$& 0.6305& 0.4956&0.4269\\
 $\mathbf{SearchLLM}$& \textbf{0.7744}& \textbf{0.7230}&\textbf{0.4435}\\\hline
 $\mathbf{Binoculars}$
& 0.2238& 0.8676&0.5115\\
 $\mathbf{SearchLLM}$& \textbf{0.6121}& \textbf{0.9255}&\textbf{0.5419}\\

    \end{tabular}
    \caption{Detection of LLM-generated text manipulated by the $\mathrm{DIPPER}$ attack.}
    \label{tab:DIPPER_attack}
\end{table}


The results demonstrate that $\mathrm{DIPPER}$ is capable of mounting successful attacks against all existing methods under evaluation, highlighting a significant vulnerability in current approaches. Notably, the majority of methods experience a remarkable degradation in performance, with  a ROC AUC rates dropped when subjected to the $\mathrm{DIPPER}$ attack. However, $\mathrm{SearchLLM}$ offers a degree of resilience, particularly in scenarios involving LLM-targeted attacks, as it can effectively mitigate the adverse effects imposed by $\mathrm{DIPPER}$. 

\subsection{Google Search Scenario}
\label{sec:google_search}


To evaluate detection performance beyond Wikipedia-based texts, we conduct experiments using news-related and 
QA-related content from the MAGE dataset~\cite{li2024mage}.
Other domains from the RAID dataset~\cite{dugan2024raid} and low-resource languages from the M4 dataset~\cite{wang2024m4} are presented in Appendix~\ref{appendix:section:other_domain_language}.
Specifically, we leverage the Google Search API to facilitate web-scale analysis but, given the associated high costs, we randomly sample 300 human-written and GPT-3.5-turbo-generated texts from the MAGE dataset. This sample size aligns with the experimental setup employed in the $\mathrm{DetectGPT}$ paper~\cite{mitchell2023detectgpt}. 

Furthermore, because the $\mathrm{Longformer}$ detector~\cite{li2024mage} is trained on MAGE data, we extend our evaluation by randomly selecting 150 human-written articles from the XSum dataset~\cite{narayan2018don}. We then generate paraphrased versions of these articles using GPT-4o-mini (4o-mini) and GPT-4o (4o). Additionally, we experiment with revised and polished texts, as shown in Table~\ref{tab:google_search}. 

The experimental results demonstrate that existing methods perform strongly on the MAGE dataset. However, all existing methods exhibit a significant decline in performance on the XSum dataset, where the LLM-generated text closely mimics human text. $\mathrm{SearchLLM}$ improves upon existing methods, particularly with close-mimic texts from the XSum dataset. Moreover, $\mathrm{SearchLLM}$ produces more gaps with other methods on both MAGE and XSum dataset for detecting short texts as illustrated in Appendix~\ref{sec:appendix_google_short_text}.


\section{Conclusion}

In this paper, we propose a novel method, $\mathrm{SearchLLM}$, for the detection of texts paraphrased by LLMs. $\mathrm{SearchLLM}$ operates by first locating potential source material on the internet using a search engine, thereby establishing a pool of candidate texts for comparison. It then assesses the similarity between the input text and a re-generated version derived from these candidate sources to accurately identify LLM-paraphrased content. Experimental results indicate that $\mathrm{SearchLLM}$ serves effectively as a proxy to enhance the performance of existing LLM detection methods. Furthermore, our findings suggest that $\mathrm{SearchLLM}$ is robust against attacks, particularly those involving paraphrasing techniques intended to evade detection, thereby addressing a key vulnerability in current approaches.

\section*{Acknowledgments}

We would like to express our sincere gratitude to the anonymous reviewers and area chairs for their valuable comments.

\section*{Limitations}

\paragraph{Private Source}
A principal limitation of $\mathrm{SearchLLM}$ lies in its reliance on external search engines for information retrieval. Consequently, its scope is inherently restricted to public sources, specifically those that are indexed and accessible via these search engines. This dependency means that $\mathrm{SearchLLM}$ is unable to process or retrieve information from private data sources, such as confidential internal documents or personal social media posts authored by individuals.

\paragraph{Freely LLM Text}
$\mathrm{SearchLLM}$ demonstrates limitations when applied to texts that are freely generated by LLMs without being directly based on identifiable sources.
Furthermore, when the output of an LLM diverges significantly in meaning or structure from any original source, or when it synthesizes information from sources in a different language, $\mathrm{SearchLLM}$ struggles to trace the origins of the content.

\paragraph{Cost}
$\mathrm{SearchLLM}$ utilizes an search engine to retrieve information from the internet and an LLM to  regenerate text. Utilizing commercial platforms, such as Google Search or proprietary LLMs like ChatGPT, can incur substantial expenses due to access fees and API usage costs. To mitigate these financial burdens, alternative approaches leverage local or free search engines, such as Wikipedia, and open-source LLMs. While these options typically offer reduced operational costs, they are often constrained by limitations in domain coverage and model capability. Thus, a balance must be struck between cost-effectiveness and the breadth and quality of information retrieval and generation when designing such systems.
In particular, for information retrieval, we utilize the Wikipedia API\footnote{\url{https://pypi.org/project/wikipedia/}} free of charge for queries related to Wikipedia content. For other types of text, $\mathrm{SearchLLM}$ issues one query per sample using the Google Search API\footnote{\url{https://developers.google.com/custom-search/v1/overview}}, which incurs a cost of \$0.005 per query. Regeneration is performed only when a suitable candidate is identified. Notably, as shown in Figure~\ref{fig:breakdown_performance_gpt_4o_mini}, approximately 50\% of texts yield a valid candidate. With an average of 378.5 tokens per sample, the regeneration cost per sample remains below \$0.00015 from GPT-4o-mini API (\$0.00003 for 200 input tokens and \$0.00012 for 200 output tokens, respectively). Thus, on average, $\mathrm{SearchLLM}$ consumes less than \$0.00515 per sample. This is lower than the cost of any GPT models with web search\footnote{\url{https://openai.com/api/pricing/}}, which are priced at \$0.01 per query, as well as other popular models with web search, such as Claude models\footnote{\url{https://www.claude.com/pricing\#api}}  (\$0.01 per query) and Gemini Flash models\footnote{\url{https://ai.google.dev/gemini-api/docs/pricing}} (\$0.035 per query).

\paragraph{Adaptive Attack}

The robustness of $\mathrm{SearchLLM}$ can be compromised if an attacker intentionally disseminates manipulated text across the internet with the aim of having it indexed by major search engines. This strategy allows the attacker to indirectly inject malicious or misleading content into the retrieval-augmented pipeline of $\mathrm{SearchLLM}$, thereby influencing the model’s outputs. 
To mitigate this vulnerability, it is advisable to constrain search queries to trustworthy sources, such as established news organizations like BBC News.
Furthermore, restricting retrieval to textual content published prior to the emergence of LLMs adds an additional safeguard.

\bibliography{1_SeachLLM}

\appendix

\section{Algorithms}

The processes of $\mathrm{SearchLLM}$ for detecting LLM-generated text and matching input sentences with URL content are described in Algorithm~\ref{alg:LLM_text_detection} and Algorithm~\ref{alg:matching}, respectively.

\label{appendix:session:algorithm}
\newcommand\mycommfont[1]{\small\ttfamily\textcolor{black}{#1}}
\SetCommentSty{mycommfont}

\begin{algorithm*}[!t]
\caption{$\mathrm{SearchLLM}$ for LLM text detection.} 
\label{alg:LLM_text_detection}

\SetKwInOut{KwIn}{Input}
\SetKwInOut{KwOut}{Output}

\KwIn{Input text $t$}
\KwOut{$Original/Generated$, $Confidence$ $score$}

    $U \leftarrow {\mathrm{SEARCH}}(t)$ \Comment{List of urls}
    
    \For{each $u_i$ in $U$}{
        $t_c \leftarrow {\mathrm{EXTRACT\_CANDIDATE}}(u_i,t)$
        
        \If{ $t_c \ne None$ }{
            $\sigma_c \leftarrow {\mathrm{MEASURE\_SIMILARITY}}(t_c,t)$

            \uIf {$\sigma_c \ge  \alpha$ }{
                \KwRet{$Original$, $\sigma_c$}                   
            }
            \Else{
                $t_r \leftarrow {\mathrm{REGENERATE}}(t_c)$         

                $\sigma_r \leftarrow {\mathrm{MEASURE\_SIMILARITY}}(t_r,t)$

                \uIf {$\sigma_r - \sigma_c \ge  \Delta$ }{
                    \KwRet{$Generated$, $\sigma_r - \sigma_c$}                   
                }
                \Else{
                    \KwRet{${\mathrm{DETECT\_BY\_AN\_EXISTING\_METHOD}}(t)$}                   
                }                                    
            }
        }
    }
\KwRet{${\mathrm{DETECT\_BY\_AN\_EXISTING\_METHOD}}(t)$}                   

\end{algorithm*}

\begin{algorithm*}[!t]
\caption{Matching input sentences with URL content.} 
\label{alg:matching}

\SetKwInOut{KwIn}{Input}
\SetKwInOut{KwOut}{Output}

\KwIn{Input text $t$, URL $u$}
\KwOut{Matched pair $P$}

    $S_t \leftarrow {\mathrm{SPLIT\_SENTENCE}}(t)$ 
    
    $S_u \leftarrow {\mathrm{SPLIT\_SENTENCE}}(u)$

    $b \leftarrow \emptyset$ \Comment{Best index}

    \For{each $S_{u_i}$ in $S_u$}{
        \If{$b == \emptyset$ or $\mathrm{SIM}(S_{t_1}, S_{u_i}) > \mathrm{SIM}(S_{t_1}, S_{u_b})$}{
            $b \leftarrow i$
        }
    }

    $S_u \gets S_u \setminus \{s_1, s_2, \ldots, s_{b-1}\}  $ \Comment{Update $S_u$ by removing its first $(b-1)$ sentences.}

    $P \leftarrow \emptyset$ \Comment{Matched pairs}
    
    \While{$S_u \ne \emptyset$ and $S_t \ne \emptyset$}{

        $p \leftarrow \emptyset$ \Comment{Best current pair}

        \For{$k$ from $1$ to $len(S_u)$}{

            $C_k = S_u|_{1:k}   $ \Comment{the first $k$ sentences of $S_u$}

            \If{$p = 0$ or $\mathrm{SIM}(S_{t_1},C_{k}) > \mathrm{SIM}(p)$}{
                $p \leftarrow \{S_{t_1},C_k\} $             
            }
        }
        
        \For{$l$ from $1$ to $len(S_t)$}{

            $C_l = S_t|_{1:l}   $ \Comment{the first $l$ sentences of $S_t$}

            \If{$\mathrm{SIM}(C_{l},S_{u_1}) > \mathrm{SIM}(p)$}{
                $p \leftarrow \{C_l,S_{u_1}\} $             
            }
        }

        $P \gets P \cup \{p\}$ \Comment{Add $p$ to $P$}

        $S_t \leftarrow S_t \setminus p$ \Comment{Update $S_t$ by removing $p$ from $S_t$}

        $S_u \leftarrow S_u \setminus p$ \Comment{Update $S_u$ by removing $p$ from $S_u$}        

    }

    \KwRet{$P$}                   
    
\end{algorithm*}

\section{Other Metrics}
\label{appendix:other_metrics}

As shown in Table~\ref{tab:RAID_wikipedia_F1}, we report the $F$-score from the same experiments described in Table~\ref{tab:RAID_wikipedia_ROC_AUC}. The results demonstrate that the $F$-score yields outcomes similar to the ROC AUC score. 
We also report the ROC AUC at an FPR of 1\% for other main tables in this paper, as shown in Tables~\ref{tab:RAID_paraphrasing_ROC_AUC_FPT_1}--\ref{tab:google_search_ROC_AUC_FPT_1}. $\mathrm{SearchLLM}$ demonstrates more stable scores than $\mathrm{Binoculars}$ and achieves a higher average score overall. In the other tables, $\mathrm{SearchLLM}$ achieves ROC AUC scores that are greater than or equal to those of other existing methods.


\begin{table*}[!t]
    \centering
    \setlength\tabcolsep{5.0pt} 
    \begin{tabular}{l c c c c c }  
         \textbf{Method}&  \textbf{GPT-4}&  \textbf{Llama-2-70B-Chat}&  \textbf{Cohere-Chat}&  \textbf{MPT-30B-Chat}& \textbf{Mistral-7B-Chat}\\ \hline  
         $\mathbf{Longformer}$&  0.9705&  0.9879&  0.8827&  0.9609& 0.9592\\ 
 $\mathbf{SearchLLM}$& \textbf{0.9802}& \textbf{0.9943}&\textbf{0.9195}&  \textbf{0.9768}&\textbf{0.9731}\\\hline
 $\mathbf{RADAR}$& 0.9605& 0.9794& 0.8958& 0.9759&0.9712\\
 $\mathbf{SearchLLM}$& \textbf{0.9702}& \textbf{0.9866}& \textbf{0.9249}& \textbf{0.9832}&\textbf{0.9796}\\\hline
$\mathbf{DetectGPT}$& 0.8298& 0.8683& 0.7383& 0.8031&0.8305\\
 $\mathbf{SearchLLM}$& \textbf{0.8867}& \textbf{0.9143}& \textbf{0.8418}& \textbf{0.8781}& \textbf{0.8856}\\\hline
 $\mathbf{Binoculars}$
& 0.9912& \textbf{0.9997}& 0.9847&\textbf{0.9997}&0.9985\\
 $\mathbf{SearchLLM}$& \textbf{0.9935}&\textbf{0.9997}& \textbf{0.9893}& \textbf{0.9997}& \textbf{0.9985}\\

    \end{tabular}
    \caption{Performance of LLM-generated text detection on all Wikipedia-related samples from the RAID dataset (F-score).}
    \label{tab:RAID_wikipedia_F1}
\end{table*}



\begin{table*}[!t]
    \centering
    \begin{tabular}{l c c c c c }  
         \textbf{Method}&    \textbf{GPT-4o-mini}
&\textbf{GPT-4.1}&  \textbf{DeepSeek-V3}&  \textbf{Grok-3-mini}& \textbf{Phi-4}\\ \hline 
         $\mathbf{Longformer}$&    0.0076&0.0099&  0.0024&  0.0015&  0.0025\\   
 $\mathbf{SearchLLM}$&   \textbf{0.4231}& \textbf{0.3977}& \textbf{0.3930}& \textbf{0.3079}& \textbf{0.3530}\\\hline
 $\mathbf{RADAR}$&   0.0136&0.0132&  0.0229& 0.0217& 0.0184\\
 $\mathbf{SearchLLM}$&   \textbf{0.4173}&\textbf{0.3594}& \textbf{0.4178}& \textbf{0.3225}& \textbf{0.3660}\\\hline
 $\mathbf{DetectGPT}$&   0.0065& 0.0062& 0.0049& 0.0092& 0.0046 \\
 $\mathbf{SearchLLM}$&   \textbf{0.4122}&\textbf{0.3820}& \textbf{0.3879}& \textbf{0.3467}& \textbf{0.3354}\\\hline
 $\mathbf{Binoculars}$&   0.0115&0.0168& 0.0090& 0.0113& 0.0039\\
 $\mathbf{SearchLLM}$& \textbf{0.4188}&\textbf{0.3890}& \textbf{0.3989}&\textbf{0.3592}& \textbf{0.3321}\\

    \end{tabular}
    \caption{Detection of paraphrased text generated by various large language models (ROC AUC at an FPR of 1\%).}
    \label{tab:RAID_paraphrasing_ROC_AUC_FPT_1}
\end{table*}


\begin{table*}[!t]
    \centering
    \begin{tabular}{ l  c c c c c c c c}
\multirow{2}{*}{\textbf{Generation}} & $\mathbf{Para}$& $\mathbf{Para}$ & $\mathbf{Para}$ 
& $\mathbf{Para}$ 
&$\mathbf{Para}$ 
& $\mathbf{Revise}$ &$\mathbf{Revise}$ &$\mathbf{Polish}$\\
 & \textbf{4o-mini}& \textbf{4.1} & \textbf{DeepSeek}& \textbf{Grok}&\textbf{Phi}& \textbf{4o-mini} &\textbf{4.1} &\textbf{4o-mini}\\\hline
 \multirow{2}{*}{\textbf{Regeneration}}& $\mathbf{Para}$& $\mathbf{Para}$ & $\mathbf{Para}$ 
& $\mathbf{Para}$ 
&$\mathbf{Para}$ 
& $\mathbf{Para}$ &$\mathbf{Para}$ &$\mathbf{Para}$\\
 & \textbf{4o-mini}& \textbf{4o-mini} & \textbf{4o-mini} 
& \textbf{4o-mini} 
&\textbf{4o-mini} 
& \textbf{4o-mini} &\textbf{4o-mini} &\textbf{4o-mini}\\ \hline 
         $\mathbf{Longformer}$&  0.0076&   0.0099& 0.0024& 0.0015&0.0025&  0.0089&0.0058& 0.0048\\   
 $\mathbf{SearchLLM}$& \textbf{0.4231}&   \textbf{0.2981}& \textbf{0.2649}& \textbf{0.2603}&  \textbf{0.2603}& \textbf{0.3355}& \textbf{0.1947}& \textbf{0.0957}\\ \hline
 $\mathbf{RADAR}$& 0.0136&  0.0132&  0.0229&  0.0217&0.0184& 0.0107& 0.0137&0.0150\\ 
 $\mathbf{SearchLLM}$& \textbf{0.4173}&  \textbf{0.3087}& \textbf{0.2906}& \textbf{0.2848}&\textbf{0.2834}& \textbf{0.3555}&\textbf{0.2314}&\textbf{0.1172}\\\hline
 $\mathbf{DetectGPT}$& 0.0065&  0.0088& 0.0088& 0.0044&0.0044&  0.0044& 0.0044&0.0133\\
 $\mathbf{SearchLLM}$& \textbf{0.4122}&   \textbf{0.8542}& \textbf{0.2044}& \textbf{0.1688}& \textbf{0.1911}&   \textbf{0.2977}& \textbf{0.1911}&\textbf{0.0800}\\\hline
 $\mathbf{Binoculars}$& 0.0115& 0.0168& 0.0090& 0.0113&0.0039&  0.0228&0.0251&0.0221\\
 $\mathbf{SearchLLM}$&  \textbf{0.4188}& \textbf{0.3369}& \textbf{0.2545}& \textbf{0.2670}&\textbf{0.2604}& \textbf{0.4137}&\textbf{0.2526}&\textbf{0.1329}\\
    \end{tabular}
    \caption{Detection of LLM-generated text when the LLMs and prompts are unknown (ROC AUC at an FPR of 1\%).}
    \label{tab:unknown_model_prompt_ROC_AUC_FPT_1}
\end{table*}


\begin{table}[!t]
    \centering
        \setlength\tabcolsep{6.0pt} 
    \begin{tabular}{l c c c }  
         \textbf{Method}&  \textbf{Human} &  \textbf{LLM}&  \textbf{Both}\\ \hline 
         $\mathbf{Longformer}$&  0.0075&  0.0064&  0.0051\\   
 $\mathbf{SearchLLM}$& \textbf{0.3723}& \textbf{0.0687}& \textbf{0.0087}\\ \hline 
 $\mathbf{RADAR}$& 0.0000& 0.0376& 0.0032\\ 
 $\mathbf{SearchLLM}$& \textbf{0.3723}& \textbf{0.1043}& \textbf{0.0087}\\\hline
 $\mathbf{DetectGPT}$& 0.0177& \textbf{0.0000}&0.0044\\
 $\mathbf{SearchLLM}$& \textbf{0.3155}& \textbf{0.0000}&\textbf{0.0311}\\\hline
 $\mathbf{Binoculars}$ & 0.0008& 0.1504&0.0037\\
 $\mathbf{SearchLLM}$& \textbf{0.3723}& \textbf{0.2565}&\textbf{0.0087}\\

    \end{tabular}
    \caption{Detection of LLM-generated text manipulated by the $\mathrm{DIPPER}$ attack (ROC AUC at an FPR of 1\%).}
    \label{tab:DIPPER_attack_ROC_AUC_FPT_1}
\end{table}


\begin{table*}[!t]
    \centering
\setlength\tabcolsep{1.5pt} 
    \begin{tabular}{ l | c  c | c c c c }  
         \textbf{Dataset}&  \textbf{MAGE (News)}&\textbf{MAGE (QA)}&  \multicolumn{4}{c}{\textbf{XSum}}\\\hline
\multirow{2}{*}{\textbf{Generation}} & $\mathbf{Topic}$-$\mathbf{based}$ &$\mathbf{Topic}$-$\mathbf{based}$ & $\mathbf{Paraphrase}$ & $\mathbf{Paraphrase}$& $\mathbf{Revise}$&$\mathbf{Polish}$\\
 & \textbf{3.5-turbo} &\textbf{3.5-turbo} & \textbf{4o-mini}& \textbf{4o} & \textbf{4o-mini}&\textbf{4o-mini}\\
 \multirow{2}{*}{\textbf{Regeneration}}& $\mathbf{Paraphrase}$ &$\mathbf{Paraphrase}$ & $\mathbf{Paraphrase}$& $\mathbf{Paraphrase}$& $\mathbf{Paraphrase}$&$\mathbf{Paraphrase}$\\
 & \textbf{4o-mini} &\textbf{4o-mini} & \textbf{4o-mini}&  \textbf{4o-mini}& \textbf{4o-mini}&\textbf{4o-mini}\\ \hline 
         $\mathbf{Longformer}$& \textbf{0.0000}&0.0466& 0.0155& 0.0155&0.0222&0.0244\\   
 $\mathbf{SearchLLM}$& \textbf{0.0000}&\textbf{0.3822}& \textbf{0.1177}& \textbf{0.1177}&\textbf{0.1644}& \textbf{0.0822}\\ \hline 
 $\mathbf{RADAR}$& \textbf{0.0000}&\textbf{0.0000}& \textbf{0.0000}& \textbf{0.0000}&0.0088&0.0088\\ 
 $\mathbf{SearchLLM}$& \textbf{0.0000}&\textbf{0.0000}& \textbf{0.0000}& \textbf{ 0.0000}&\textbf{0.1600}&\textbf{0.0666}\\\hline
 $\mathbf{DetectGPT}$& \textbf{0.0000}&\textbf{0.2088}& \textbf{0.0000}& \textbf{0.0000}&\textbf{0.0000}&\textbf{0.0000}\\
 $\mathbf{SearchLLM}$& \textbf{0.0000}&\textbf{0.2088}& \textbf{0.0000}& \textbf{0.0000}&\textbf{0.0000}&\textbf{0.0000}\\\hline
 $\mathbf{Binoculars}$ & 0.6488& \textbf{0.6533}& 0.0044& 0.0044&0.0533&0.0266\\
 $\mathbf{SearchLLM}$&  \textbf{0.6577}& \textbf{0.6533}& \textbf{0.1066}& \textbf{0.1066}&\textbf{0.2044}&\textbf{0.0844}\\
    \end{tabular}
    \caption{Detection of LLM generated text using Google Search Engine (ROC AUC at an FPR of 1\%).}
    \label{tab:google_search_ROC_AUC_FPT_1}
\end{table*}

\section{Statistical Test}
\label{appendix:section:statisical_test}

In accordance with the experiment presented in Table~\ref{tab:RAID_paraphrasing}, we conduct paired $t$-tests to compare the performance of existing methods with and without the support of $\mathrm{SearchLLM}$. The results, shown in Table~\ref{tab:statistical_test}, demonstrate that all methods supported by $\mathrm{SearchLLM}$ perform significantly better than those without support, with all $p$-values less than $0.004$.

\begin{table*}[!t]
    \centering
\setlength\tabcolsep{6pt} 
    \begin{tabular}{l r r r r r }  
         \textbf{Method}&    \textbf{GPT-4o-mini}
&\textbf{GPT-4.1}&  \textbf{DeepSeek-V3}&  \textbf{Grok-3-mini}& \textbf{Phi-4}\\ \hline 
         $\mathbf{SearchLLM}$ vs $\mathbf{Longformer}$ &    < 0.0001 &< 0.0001&  < 0.0001&  < 0.0001&  < 0.0001\\
 $\mathbf{SearchLLM}$ vs $\mathbf{RADAR}$&   0.0033&< 0.0001& < 0.0001& < 0.0001& < 0.0001\\
 $\mathbf{SearchLLM}$ vs $\mathbf{DetectGPT}$&   
< 0.0001&< 0.0001& < 0.0001& < 0.0001& < 0.0001\\
 $\mathbf{SearchLLM}$ vs $\mathbf{Binoculars}$&   0.0020&< 0.0001& < 0.0001& < 0.0001& < 0.0001\\

    \end{tabular}
    \caption{$p$-values from paired $t$-tests assessing the detection of LLM-paraphrased text generated by various language models.}
    \label{tab:statistical_test}
\end{table*}

\section{Breakdown of Performance on Detecting GPT-4.1 Generated Text}
\label{sec:appendix:breakdown_gpt_4_1}

\begin{figure}[!t]
    \centering
    \includegraphics[width=1\linewidth]{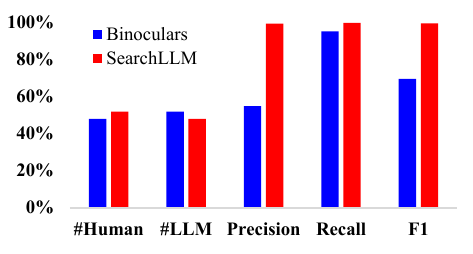}
    \caption{The number of human- and LLM-generated texts processed by $\mathrm{SearchLLM}$ and $\mathrm{Binoculars}$, along with their performance in detecting GPT-4.1-generated content.}
    \label{fig:break_down_gpt-4-1}
\end{figure}

We provide a detailed analysis of the performance of $\mathrm{SearchLLM}$ combined with $\mathrm{Binoculars}$ in detecting GPT-4.1 generated text, as shown in Figure~\ref{fig:break_down_gpt-4-1}. 
$\mathrm{SearchLLM}$ and $\mathrm{Binoculars}$ demonstrate a similar trend when detecting GPT-4o-mini generated text, as illustrated in Figure~\ref{fig:breakdown_performance_gpt_4o_mini}.



\section{Change of Other Parameters}
\label{sec:appendix:other_parameters}

\begin{figure}[!t]
    \centering
    \includegraphics[width=1\linewidth]{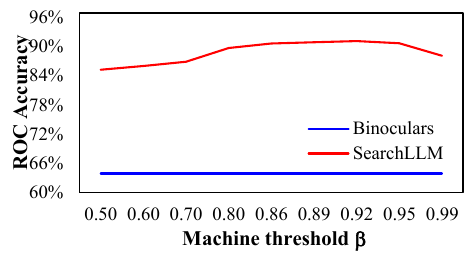}
    \caption{The impact of varying the machine threshold parameter $\beta$ on the performance of $\mathrm{SearchLLM}$.}
    \label{fig:paraphrase_threshold}
\end{figure}

\begin{figure}[!t]
    \centering
    \includegraphics[width=1\linewidth]{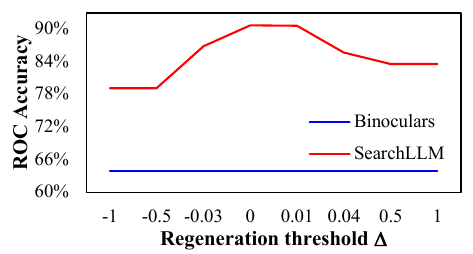}
    \caption{The impact of varying the regeneration threshold parameter $\Delta$ on the performance of $\mathrm{SearchLLM}$.}
    \label{fig:regeneration_threshold}
\end{figure}

\begin{figure}[!t]
    \centering
    \includegraphics[width=1\linewidth]{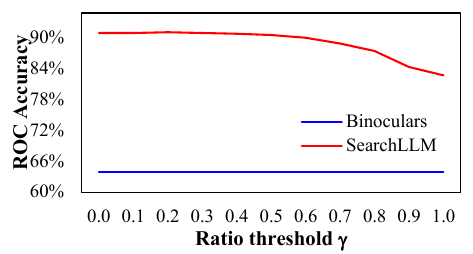}
    \caption{The impact of varying the ratio threshold parameter $\gamma$ on the performance of $\mathrm{SearchLLM}$.}
    \label{fig:ratio_threshold}
\end{figure}

In addition to the human threshold $\alpha$ discussed in Figure~\ref{fig:human_threshold_alpha}, we also evaluate the effects of varying other thresholds, 
including the machine threshold $\beta$ (Figure~\ref{fig:paraphrase_threshold}), the regeneration threshold $\Delta$ (Figure~\ref{fig:regeneration_threshold}), and the ratio threshold $\gamma$ (Figure~\ref{fig:ratio_threshold}). 
The results indicate that the optimal values for $\beta$, $\Delta$, and $\gamma$ are relatively high, low, and middle ranges, respectively.



\section{Various Temperature Settings}
\label{appendix:section:temperature}

\begin{figure}[!t]
    \centering
    \includegraphics[width=1\linewidth]{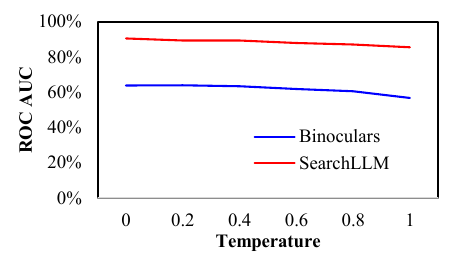}
    \caption{Performance comparison of $\mathrm{Binoculars}$ and $\mathrm{SearchLLM}$ in detecting LLM-generated text across different temperature settings.}
    \label{fig:tempreture_binoculars}
\end{figure}

\begin{figure}[!t]
    \centering
    \includegraphics[width=1\linewidth]{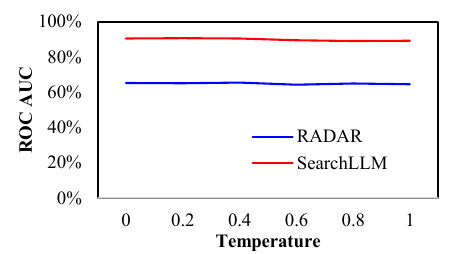}
    \caption{Performance comparison of $\mathrm{RADAR}$ and $\mathrm{SearchLLM}$ in detecting LLM-generated text across different temperature settings.}
    \label{fig:tempreture_RADAR}
\end{figure}

We conduct experiments to evaluate the detection of LLM-paraphrased text generated by GPT-4o-mini at various temperature settings, as shown in Figure~\ref{fig:tempreture_binoculars} and Figure~\ref{fig:tempreture_RADAR}. The results indicate that $\mathrm{Binoculars}$ is more sensitive to temperature changes than $\mathrm{RADAR}$. Meanwhile, $\mathrm{SearchLLM}$ consistently demonstrates improved performance across different temperatures.


\section{Composite}
\label{appendix:section:composite}

We evaluate the ability to detect composite text by dividing each sample into equal parts. The first part is either written by a human or paraphrased by GPT-4o-mini, while the second part is paraphrased or revised by GPT-4o-mini. The results, shown in Table~\ref{tab:composite}, demonstrate that $\mathrm{SearchLLM}$ outperforms existing methods across all combinations.

\begin{table*}[!t]
    \centering
        \setlength\tabcolsep{6.0pt} 
    \begin{tabular}{l c c c }  
         \textbf{Method}&  \textbf{Human and Paraphrase} &  \textbf{Human and Revise}&  \textbf{Paraphrase and Revise}\\ \hline 
         $\mathbf{Longformer}$&  0.5660&  0.5850&  0.6316\\   
 $\mathbf{SearchLLM}$& \textbf{0.7866}& \textbf{0.7907}& \textbf{0.8870}\\ \hline 
 $\mathbf{RADAR}$& 0.5846& 0.6082& 0.6671\\ 
 $\mathbf{SearchLLM}$& \textbf{0.7781}& \textbf{0.7800}& \textbf{0.8984}\\\hline
 $\mathbf{DetectGPT}$& 0.5499&  0.5816&0.6109\\
 $\mathbf{SearchLLM}$& \textbf{0.7672}& \textbf{0.7694}&\textbf{0.8723}\\\hline
 $\mathbf{Binoculars}$&  0.5667& 0.5811&0.6389\\
 $\mathbf{SearchLLM}$& \textbf{0.7862}& \textbf{0.7821}&\textbf{0.8866}\\

    \end{tabular}
    \caption{Detection of composite text generated by large language models.}
    \label{tab:composite}
\end{table*}

\section{Google Search Scenario with Other Datasets}
\label{appendix:section:other_domain_language}

We evaluate the capability of our method in detecting all other domains within the RAID dataset~\cite{dugan2024raid}, including News, Book, Abstract, Poetry, Recipe, Reddit, and Reviews, as shown in Table~\ref{tab:other_domain_RAID}. $\mathrm{SearchLLM}$ consistently achieves a ROC AUC greater than 0.8 across all domains, outperforming $\mathrm{DetectGPT}$ and matching or exceeding the performance of other existing methods.



\begin{table*}[!t]
    \centering
    \begin{tabular}{ l   c  c c c c c c  }
 \textbf{Method}& \textbf{News} & \textbf{Books}& \textbf{Abstract}& \textbf{Poetry}&\textbf{Recipe} &\textbf{Reddit}&\textbf{Reviews}\\ \hline 
         $\mathbf{Longformer}$&  0.9934& 0.9987& \textbf{1.0000}& 0.9588& 0.8332&0.9914&\textbf{1.0000}\\   
 $\mathbf{SearchLLM}$&  \textbf{0.9999}& \textbf{0.9994}& \textbf{1.0000}& \textbf{0.9817}& \textbf{0.8499}&\textbf{0.9955}&\textbf{1.0000}\\ \hline 
 $\mathbf{RADAR}$&  \textbf{1.0000}& 0.9995& 0.9946& 0.8462& 0.9973&0.9841&0.9948\\ 
 $\mathbf{SearchLLM}$&  \textbf{1.0000}& \textbf{0.9996}& \textbf{1.0000}& \textbf{0.9078}& \textbf{0.9973}& \textbf{0.9942}& \textbf{1.0000}\\\hline
 $\mathbf{DetectGPT}$&  0.7861& 0.9292& 0.8830& 0.7475& 0.9519&0.9561&0.9387\\
 $\mathbf{SearchLLM}$&   \textbf{0.9180}& \textbf{0.9825}& \textbf{0.9956}& \textbf{0.9451}& \textbf{0.9624}&\textbf{0.9666}&\textbf{0.9721}\\\hline
 $\mathbf{Binoculars}$ &  \textbf{1.0000}& \textbf{1.0000}&\textbf{1.0000}& 0.9999& \textbf{0.9999}&0.9984& 0.9999\\
 $\mathbf{SearchLLM}$&  \textbf{1.0000}& \textbf{1.0000}& \textbf{1.0000}& \textbf{1.0000}& \textbf{0.9999}&\textbf{0.9992}&\textbf{1.0000}\\
    \end{tabular}
    \caption{Detection of LLM-generated text across all other domains in the RAID dataset using the Google Search Engine.}
    \label{tab:other_domain_RAID}
\end{table*}

Furthermore, we extend our experiments to detect content in all low- and medium-resource languages in the M4 dataset~\cite{wang2024m4}, including Urdu (low-resource), Bulgarian (low-resource), and Indonesian (medium-resource), as presented in Table~\ref{tab:low_resource_languages}. The results demonstrate that $\mathrm{SearchLLM}$ efficiently improves the performance compared to all existing methods.


\begin{table*}[!t]
    \centering
    \begin{tabular}{ l   c c c }
 \textbf{Method}& \textbf{Urdu}& \textbf{Bulgarian}& \textbf{Indonesian}\\ \hline 
         $\mathbf{Longformer}$& 0.4393& 0.7703&0.7323\\   
 $\mathbf{SearchLLM}$& \textbf{0.5817}& \textbf{0.8829}&  \textbf{0.8161}\\ \hline 
 $\mathbf{RADAR}$& 0.4858& 0.6715&0.6855\\ 
 $\mathbf{SearchLLM}$& \textbf{0.6069}& \textbf{0.8459}&\textbf{0.7873}\\\hline
 $\mathbf{DetectGPT}$& 0.4371& 0.5474&0.5116\\
 $\mathbf{SearchLLM}$&  \textbf{0.5707}& \textbf{0.8122}&\textbf{0.6900}\\\hline
 $\mathbf{Binoculars}$ & 0.7968& 0.7891&0.9543\\
 $\mathbf{SearchLLM}$& \textbf{0.8386}& \textbf{0.8786}&\textbf{0.9598}\\
    \end{tabular}
    \caption{Detection of LLM-generated text in the M4 dataset using Google Search for low- and medium-resource languages.}
    \label{tab:low_resource_languages}
\end{table*}

\section{Google Search Scenario with Short Text}
\label{sec:appendix_google_short_text}

In alignment with the experiments listed in Table~\ref{tab:google_search}, we report the results of detecting LLM-generated text on short texts limited to 30 words, as shown in Table~\ref{tab:google_search_short_text}. The results demonstrate that $\mathrm{SearchLLM}$ generally exhibits larger performance gaps compared to other methods in both the MAGE and XSum experiments.



\begin{table*}[!t]
    \centering
\setlength\tabcolsep{1.5pt} 
    \begin{tabular}{ l  |c  c | c c c c }  
         \textbf{Dataset}&  \textbf{MAGE (News)}&\textbf{MAGE (QA)}&  \multicolumn{4}{c}{\textbf{XSum}}\\\hline
\multirow{2}{*}{\textbf{Generation}} & $\mathbf{Topic}$-$\mathbf{based}$ &$\mathbf{Topic}$-$\mathbf{based}$ & $\mathbf{Paraphrase}$ & $\mathbf{Paraphrase}$& $\mathbf{Revise}$&$\mathbf{Polish}$\\
 & \textbf{3.5-turbo} &\textbf{3.5-turbo} & \textbf{4o-mini}& \textbf{4o} & \textbf{4o-mini}&\textbf{4o-mini}\\
 \multirow{2}{*}{\textbf{Regeneration}}& $\mathbf{Paraphrase}$ &$\mathbf{Paraphrase}$ & $\mathbf{Paraphrase}$& $\mathbf{Paraphrase}$& $\mathbf{Paraphrase}$&$\mathbf{Paraphrase}$\\
 & \textbf{4o-mini} &\textbf{4o-mini} & \textbf{4o-mini}&  \textbf{4o-mini}& \textbf{4o-mini}&\textbf{4o-mini}\\ \hline 
         $\mathbf{Longformer}$& 0.8944&0.9345& 0.4270& 0.4505&0.4496&0.4840\\   
 $\mathbf{SearchLLM}$& \textbf{0.9348}&\textbf{0.9607}& \textbf{0.7197}& \textbf{0.7369}& \textbf{0.7378}&\textbf{0.6934}\\ \hline 
 $\mathbf{RADAR}$& 0.7879&0.5624& 0.6198& 0.6387&0.5983&0.5974\\ 
 $\mathbf{SearchLLM}$& \textbf{0.8651}& \textbf{0.6950}& \textbf{0.8099}& \textbf{0.8234}&\textbf{0.7998}&\textbf{0.7461}\\\hline
 $\mathbf{DetectGPT}$& 0.6760&0.8204& 0.4431& 0.4254
&0.4429&0.4776
\\
 $\mathbf{SearchLLM}$& \textbf{0.8136}&\textbf{0.8915}&  \textbf{0.7208}& \textbf{0.7180}&\textbf{0.7218}&\textbf{0.6739}\\\hline
 $\mathbf{Binoculars}$
& 0.8395&0.9570& 0.5707& 0.5401&0.6009&0.5785\\
 $\mathbf{SearchLLM}$& \textbf{0.9234}&\textbf{0.9747}& \textbf{0.7875}& \textbf{0.7656}&\textbf{0.8052}& \textbf{0.7216}\\
    \end{tabular}
    \caption{Detection of LLM-generated short text using the google search engine.}
    \label{tab:google_search_short_text}
\end{table*}

\end{document}